\documentclass[12pt]{article}

\usepackage{cite}

\usepackage{fix-cm}

\usepackage{dirtytalk}

\newcommand{\chapquote}[3]{\begin{flushleft}\begin{quotation} \textit{#1} \end{quotation} \end{flushleft}\begin{flushright} - #2, \textit{#3}\end{flushright} }

\usepackage{epigraph}

\usepackage{hyperref}       

\usepackage{algorithm}     
\usepackage{algpseudocode}

\title{Goals and the Structure of Experience}

\algrenewcommand\algorithmicrequire{\textbf{Input:}}
\algrenewcommand\algorithmicensure{\textbf{Output:}}

\usepackage{amsthm}
\usepackage{amsmath}
\usepackage{amsfonts}
\usepackage{graphicx}

\theoremstyle{definition}
\newtheorem*{definition*}{Definition}

\begin{document}

\maketitle

\textit{} \newcommand{\keywords}[1]{\par\addvspace\baselineskip \noindent\textbf{\textit{Keywords---}} #1}

\author{Nadav Amir$^{1}$, Stas Tiomkin$^{2}$
, and Angela Langdon$^{3}$\\
$^{1}$Princeton University
$^{2}$Texas Tech University
$^{3}$National Institute of Mental Health Intramural Research Program, National Institutes of Health\\}




\begin{abstract}
Purposeful behavior is a hallmark of natural and artificial intelligence. Its acquisition is often believed to rely on world models, comprising both descriptive (what \emph{is}) and prescriptive (what is \emph{desirable}) aspects that identify and evaluate state of affairs in the world, respectively.
Canonical computational accounts of purposeful behavior, such as reinforcement learning, posit distinct components of a world model comprising a \emph{state representation} (descriptive aspect) and a \emph{reward function} (prescriptive aspect). However, an alternative possibility, which has not yet been computationally formulated, is that these two aspects instead co-emerge interdependently from an agent's \emph{goal}. Here, we describe a computational framework of goal-directed state representation in cognitive agents, in which the descriptive and prescriptive aspects of a world model co-emerge from agent-environment interaction sequences, or \emph{experiences}. Drawing on Buddhist epistemology, we introduce a construct of goal-directed, or \emph{telic}, states, defined as classes of goal-equivalent experience distributions. Telic states provide a parsimonious account of goal-directed learning in terms of the statistical divergence between behavioral policies and desirable experience features. We review empirical and theoretical literature supporting this novel perspective and discuss its potential to provide a unified account of behavioral, phenomenological and neural dimensions of purposeful behaviors across diverse substrates.

\end{abstract}


\maketitle
\section{Introduction: goals and state representations}
\label{sec:intro}
Recent advances in cognitive science and artificial intelligence (AI) have highlighted similarities and differences between biological and artificial learning \cite{summerfield2022natural}. Arguably, the fundamental construct driving any form of intelligent behavior is that of \emph{goals}, the primary objective of purposeful behavior \cite{rosenblueth1943behavior}. While the construct of goals is a focus of (renewed) interest in cognitive science \cite{molinaro2023goal,chu2023praise,davidson2024goals,colas2024should,de2023goals}, so far only a few computational frameworks have focused directly on goals as drivers of purposeful action in cognitive agents \cite{pezzulo2009thinking,o2014goal}. Furthermore, such frameworks mostly do not explicitly address the role of goals in structuring the mental representations, or world models, that are thought to enable purposeful behaviors in the first place. Here, we describe a new computational approach accounting for the phenomenology of purposeful behavior by focusing on the reciprocal relationship between goals and the representational structures that enable their realization.  

Goal-directed behavior is often studied using reinforcement learning (RL), a canonical computational framework of trial-and-error learning, planning, and decision making in natural and artificial agents \cite{sutton2018reinforcement}. A foundational notion in RL models is that of a state representation: a compact description of the causal structure of the environment within the context of a particular task \cite{niv_learning_2019,kaelbling1998planning}. Standard RL models assumes that all goals are expressible as the maximization of value, or an expected cumulative sum of a appropriately designed scalar reward signal furnished by the environment. This assumption, known as \emph{the reward hypothesis}, lies at the heart of current RL theory and its applications \cite{bowling_settling_2023}.  A substantial subclass of RL agents learn by using state-value associations to select a course of action according to a policy that optimizes value. In practice, RL models often consider highly rewarding states as end goals and states that are associated with high expected cumulative reward, or value, as sub goals. This approach has given rise to many successful applications in robotics, control systems, AI and cognitive science, with RL becoming an influential quantitative framework for elucidating the cognitive, computational and neural underpinnings of reward-driven learning \cite{ho2022cognitive, botvinick2019reinforcement,sutton1981toward,montague1995bee, montague1996framework, niv2009reinforcement}. More recently, RL algorithms scaffolded by reward-optimized deep neural network  methods have achieved super-human level at playing video games \cite{mnih2015human}, mastered games such as Chess or Go \cite{silver2017mastering}, trained robots to execute complex manipulation skills \cite{gu2017deep} and enabled large language models to learn mathematical reasoning capabilities \cite{guo2025deepseek}. However, despite their success in  effectively learning a wide range of goal-directed behaviors \cite{silver2021reward}, recent theoretical work suggests that not all meaningful goals and tasks can be fully expressed using reward signals and value functions \cite{carr2024conditions, abel2021expressivity}. Moreover, many instances of biological learning seem more naturally explained in terms of intrinsic drives and motivations that are are not easily expressed as maximization of externally furnished scalar reward signals \cite{karayanni2022extrinsic}. A more cognitively plausible account of goal-directed learning may therefore require extending canonical reward-based RL frameworks, as successful as they may be, to formally address the construct of goals, and how they shape the phenomenology of purposeful behavior. 

The success of any RL algorithm hinges on the appropriate choice of its two central components: the reward signal and the state representation. In typical RL applications, both reward signals and state representations are specified a priori by a task designer and are selected in order to sculpt a given behavior in a specific setting. For example, training an artificial RL agent to play chess usually providing it with a state representation describing the positions of the different pieces on the chessboard, whose turn it is, whether or not the king or the rook have already moved (which determines whether one can castle) and so on. The reward signal can be simply $1$,$-1$, or $0$, depending on whether the agent wins, loses, or draws the game. For less structured tasks, such as a robot learning flexible motion skills in an unknown environment, suitable state representations for a given goal are often approximated using deep-learning or other heuristic methods\cite{lesort2018state,schaul2015universal}. In all such cases, the underlying assumption is that these representations are low-dimensional approximations of some ground-truth state-space structure. While this assumption may be useful for designing artificial systems that learn well defined tasks, in the case of biological cognitive agents there is typically no ground-truth state representation (or reward function) as these depend on the subjective goals the agent is pursuing at any moment. When trying to cross a busy road, for example, a suitable state representation should include details such as the location of the crossing and the status of the pedestrian signal. When trying to hail a taxi, on the other hand, the state representation should prioritize information about the type of incoming vehicles and their availability to accept passengers. Furthermore, while standard RL models typically include a single state representation, biological learning probably relies on multiple, hierarchically arranged ones \cite{botvinick2009hierarchically}. Finally, in any empirical setting, there is no assurance that subjects represent the task in the same way as the experimentalist \cite{song2022minimal}, or that the their preferences are aligned with those implied by the reward signal provided by the task design \cite{karayanni2022extrinsic}. Thus, traditional RL approaches that do not explicitly account for these dependencies may only partially explain how cognitive agents adaptively acquire and update state representations \cite{langdon_uncovering_2019}, and how these state representations dynamically depend on subjective goals.  

To address this gap, we have recently proposed a computational theory that focuses directly on the role of goals in shaping state representation in cognitive agents \cite{amir2023states,amir2024learning}. The theory proposes that state representations can be usefully thought of as a subjective conceptual schema for distinguishing between goal-relevant and goal-irrelevant aspects of an agent's experience. In this framework, state representations are thus not thought of as fixed and veridical models of the external world but rather fluid maps of affordances \cite{pezzulo2016navigating} that help guide the agent's behavior towards goal-aligned experiences. In this view, goals and state representations are therefore fundamentally interdependent cognitive constructs. We now turn to review the philosophical underpinnings of this proposed framework, before examining its computational and empirical implications.

\section{Philosophical foundations: the fact/value dichotomy and the origins of state representations}
\chapquote{``You cannot derive an ought from an is."}{G.E. Moore}{Principia Ethica (1903)}

The Scottish philosopher David Hume (1711-1776) famously distinguished between "is" and "ought" statements, cautioning against the tendency to infer the latter from the former \cite{sep-hume-moral}. Hume's is-ought distinction has since become a central tenet of Western ethics and epistemology, where the conflation between facts and values is considered a non sequitur, sometimes referred to as a \emph{naturalistic fallacy} \cite{moore1903principia}. The fact-value dichotomy has also influenced the development of  psychological accounts of motivation and affect by contemporary philosophers who distinguish between the cognition of facts and the cognition of values in \cite{bratman1987intention,smith1987humean}. Furthermore, it has profoundly shaped economic and decision-theoretic models of rational choice, which are grounded in a similar distinction between factual and the evaluative dimensions of choice behaviors \cite{sen1990rational}. Such models aim to explain how agents choose between different possible options based on notions of preference. Available options are construed as preconceived situations or actions, such as whether or not to accept a particular gamble or vote for a particular candidate, while preferences are expressed in terms of a utility function that provides a rank ordering of possible options \cite{von2007theory}. Such decision-theoretic frameworks were later reformulated, notably through the work of Leonard Savage and others, to incorporate subjective descriptions of the relevant contingencies, or states, that specify isolated decision making situations \cite{savage1972foundations}. Later still, RL models introduced reward signals to capture how agents' preferences over possible states can evolve by trial-and-error learning \cite{dayan2008decision}. Importantly, while state representations in RL are defined independently of reward functions, the evaluation of reward functions is typically conditioned on the current state, or the agent's best estimate thereof. This seemingly innocuous asymmetry between state representations and reward functions reflects a foundational implicit assumption about how agents model their environment. Namely, that the descriptive, or representational aspects of their world models enjoy epistemic primacy over, and independence from, the evaluative ones.  In other words, states are thought to provide veridical descriptions of situations in the world, or the agent's beliefs about such situations, regardless of any evaluation thereof. Accordingly, RL models of cognition often describe state representations as \emph{cognitive maps} that provide structural information about relationships between entities in the environment \cite{behrens2018cognitive}, which, in turn, undergird upstream evaluation and credit-assignment mechanisms. 

Alongside decision theory, another major influence on the notion of state representation in RL originates in theories of dynamical systems and control \cite{sutton2018reinforcement}. Within these theories, the \emph{state space} of a physical system describes its degrees of freedom: the smallest number of variables that need to be specified in order to fully predict or control its future behavior \cite{willems1991paradigms}. For example, the trajectory of a simple pendulum can be described by a state space comprising its angle, relative to some reference point, and angular velocity, as well as the external dynamics governing its motion, such as the forces of gravity and friction acting on it. Given initial values for these state variables, the state of the pendulum at any given timepoint can be determined. This notion of state reflects the goal of physicists and engineers, who use such state-space models to predict and control the behavior of systems they are interested in, for example, calculating the precise position of the moon in a few weeks time to safely land a spacecraft on it. 

While some capacity to predict and control the environment is crucial for the survival and well being of any adaptive organism \cite{carvalho2024predictive}, these objectives, fundamental as they may be, arguably cannot account for the full range of goal-directed behaviors exhibited by cognitive agents. 
For example, humans and other animals frequently engage in various creative or playful pursuits that cannot be easily explained solely in terms of environmental control or prediction (despite often leveraging these capacities) \cite{chu2023praise}. Such activities are more likely to be driven intrinsically by desirable affective states such as curiosity, care for others, relaxation or focused engagement \cite{csikszentmihalyi2014emerging}.
Traditional notions of state representation and reward may thus struggle to explain such behaviors since they do not view the goals of cognitive agents from the intrinsic perspective of their own idiosyncratic motivations. Rather, they view them from the perspective of an external observer with preconceived representations of state and value. Our aim here is therefore to develop an account of state representation that is sensitive to the subjective preference structures and dispositions of cognitive agents and how these shape their behavior and decision making processing. Towards this aim, we propose a new notion of state representation that unifies the descriptive and evaluative underpinnings of goal-directed behavior. 

Whereas the notion of state representation and its role in value-based learning models has been fundamentally shaped by the is/ought dichotomy in Western thought and cognitive science, Eastern philosophy of mind does not uphold such a strict dissociation between the descriptive and evaluative dimensions of cognition and mental representation. In particular, the Buddhist philosopher Dharmakīrti (circa 7th century CE), a central figure of the epistemological, or Pramāṇavāda, school of Buddhist philosophy, developed a sophisticated theory of mental representation predicated on the goal-oriented nature of concepts. Within the Buddhist ontology adopted by Dharmakīrti's, it is only particular percepts, such as the cup you see standing on your table or the sensation of heat on your hand from a nearby fire, that truly exist, while all universal concepts or categories are mental constructs that are not ultimately considered as real. 

Dharmakīrti argued that we categorize different particulars under some concept to the extent that they serve the same purpose or goal. For example, if you want to drink water, your mind may categorize different objects as ''cups"  since they can all serve the purpose of quenching your thirst \cite{dunne_foundations_2004}. Importantly, there is nothing fundamentally ``similar'' about these objects. Rather, you mentally impose a similarity on them based on their capacity to fulfill your goal. Dharmakīrti's theory thus explains how our goals, both contextual and dispositional, can shape how we represent and conceptualize our sensorimotor stream experience. 

Motivated by Dharmakīrti's theory of concept formation, we propose that a state representation can be usefully viewed as a computational operationalization of goal-directed conceptual schema. Building on this proposed analogy between states and concepts, we now turn to describe our  theory of state representation learning to explain how sensorimotor experience streams can be structured by goal-directed state representations.

\section{Theoretical setting}
\label{sec:setting}
\chapquote{``The art of being wise is the art of knowing what to overlook."}{William James}{Principles of Psychology (1890)}

In this section we outline the theoretical foundation for our proposed model. For mathematical details, see appendix \ref{appendix:formal_setting}. We begin with a definition of goals as preferences over experience distributions and use it to develop our framework of goal-directed state representation and learning.

\subsection{Goals as preferences over experience distributions}
\label{subsec:goals_as_pref}
Our approach is couched within the setting of the perception-action cycle that describes the flow of information between an agent and its environment as a stream of sensorimotor events \cite{fuster2004upper}. Using the language of RL, such streams can be defined as action-observation (or action-outcome) sequences, and we shall adopt this terminology here too. While action-observation sequences are sometimes referred to as trajectories or histories, we shall instead prefer the term \emph{experiences} to emphasize our focus on what an agent experiences as it interacts with its environment. For example, a rat navigating a maze experiences its own movements (actions), such as turning left or right at different junctions in the maze, followed by corresponding perceptual outcomes (observations), such as running into a wall or smelling a nearby food pellet. Note that experience streams are defined as temporally discrete with some relevant timescale. Interestingly, such temporal discretization of experience not only affords computational tractability, but is also supported by a recent research suggesting that conscious perception is experienced as temporally discrete \cite{herzog2020all}. 

While experiences can occur in ``real-time'', they can also be simulated or reactivated after the event, as demonstrated by phenomena such as memory reactivation \cite{wheeler2000memory} and hippocampal replay \cite{foster2017replay}. Furthermore, while an experience is defined as a single action-observation sequence, our focus here will be on the statistical regularities that emerge over multiple experiences. Thus, the construct of an \emph{experience distribution}, indicating how likely are different kinds of experiences to occur in a certain context, plays a central role in our account. Shifting the focus from properties of single experiences to those of experience distributions not only allows us to account for non-deterministic agents and environments but also nicely aligns with existing computational views on learning and cognition, such as predictive coding, or Bayesian approaches that seek to explain perception, behavior, and brain function in terms of statistical inferences over sensorimotor experiences streams \cite{knill1996perception, doya2007bayesian}. 

To formalize Dharmakīrti's theory of concept formation, we turn to define the central construct of a \emph{goal}. In line with recent theoretical investigations on the conceptual foundations of RL \cite{bowling_settling_2023}, we operationalize goals as preference relations over experience distributions. In other words, for any two experience distributions, a goal specifies whether one is preferred by the agent over the other or whether the agent is views them as equivalent. Thus, a goal can be thought of as a subjective preference, in the form of a rank ordering, over classes of experiences sharing certain statistical properties. For example, a hungry rat navigating a maze may prefer experiences, or maze trajectories, that are likely to reach locations where food is available. The same rat, when thirsty, would prefer trajectories that are more likely to reach other locations, where water is available. The goal of the rat thus consist of its current set of preferences over  possible trajectory distributions. 

From a cognitive perspective, we hypothesize that world models comprise probability distributions over possible experiences, with goals providing a preference ordering of these distributions according to their likelihood of yielding desirable outcomes, such as reaching food or water locations in the example above. What does such an account entail in terms of goal-directed learning and decision making? To answer this, we recall our definition of an experience as a sequence of actions and observations. To estimate how likely a particular experience is, the agent needs a model of its own behavior, or policy, for predicting the likelihood of generating specific actions, and a model of its environment, for predicting the likelihood of encountering different observations. Typically, agents control their own behavior but not that of their environment. Thus, we shall assume that the agent maintains a fixed model of its environment so that experience distributions are determined by the agent's policy. For the sake of simplicity, we focus here on the problem of learning goal-directed behaviors, or policies, ignoring the question of how a model of the environment is acquired or updated. 

\subsection{Telic-states as goal-equivalent experience distributions}
\label{sec:telic_states}

\chapquote{Alice: “Would you tell me, please, which way I ought to go from here?”\\
Cheshire Cat: “That depends a good deal on where you want to get to.”\\
Alice: “I don’t much care where –”\\
Cheshire Cat: “Then it doesn’t matter which way you go.”}{Lewis Carroll}{Alice in Wonderland}

While every experience is unique, learning entails generalization over experiences that are in some sense ``similar'' \cite{shepard1987toward}. The question is what makes two experiences similar, or, in short, which differences ``make a difference'' \cite{bateson1970form}? From the perspective of a learning agent, the differences that matter are precisely those specified by its goals, namely, its preferences over experience distributions. In other words, goals determine which features of experience should be taken into account, and, just as importantly, which can be ignored. Returning to the example of the hungry rat, if two (or more) policies are equally likely to yield trajectories that reach a location where food is available, they can be considered equivalent with respect to the goal of satisfying hunger. In other words, if the rat only cares about the likelihood of obtaining food, it can ignore any differences between policies that are equivalent in this regard, by clustering the resulting experience distributions into a single representational structure, or ``state''. 

\begin{figure}[htb]
    \centering
    \includegraphics[scale=0.125]{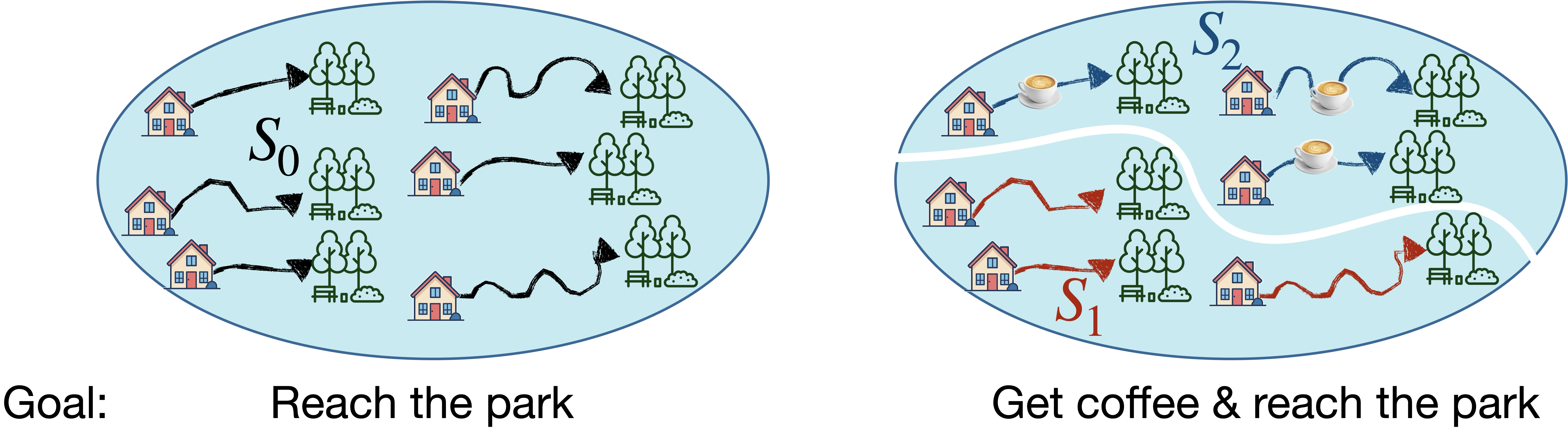}
    \caption{\textbf{Telic states are sensitive to the goal} (Left) all six possible routes, or experiences, are equally desirable with respect to the goal of reaching the park and can therefore be clustered into a single telic state $S_0$. (Right) for the goal of getting a cup of coffee on the way to the park, these experiences are split into two telic states: $S_1$, containing the routes that do not pass by the local cafe, and $S_2$ containing those that do.}
    \label{fig:telic_states}
\end{figure}

This example illustrates several core features of our approach that we can now draw together using the concept of \emph{telic states}. The word \emph{telic}, derived from the Greek \emph{telos}, meaning ultimate end, refers to something that is oriented toward a specific goal or purpose. We use it here to emphasize the explicit goal-directed aspect of state representations within our account: telic states encapsulate precisely those aspects of experience that are relevant to achieving one's goals. We can illustrate the notion of telic states using a simple example (Fig.\ref{fig:telic_states}): imagine you are planning to walk to the nearby park and are considering different possible routes for getting there from your home. If you only care about eventually reaching the park, you need not worry about which route you choose as they are all equivalent with respect to that goal. In other words, you can generalize over the (prospective) experiences of walking to the park via different routes by clustering them together into a single category, or telic state, corresponding to the goal of getting from your home to the park (Fig.\ref{fig:telic_states}, left). However, suppose that you also want to get a cup of coffee on your way to the park and that only some of the possible routes you are considering pass by your favorite café. In this case, it makes sense to split your representation of possible experiences, or routes, into two groups, namely those that stop by your café and those that don't  (Fig.\ref{fig:telic_states}, right). Thus, your goal determines how you categorize possible experiences by determining which distinctions among them are important and which can be ignored. For simplicity, we defined telic states in this example as groups of individual experiences. Generally, however, telic states consist of classes of distributions over possible experiences. More formally, each telic state can be thought of as a set of probability distributions that give the likelihood of an experience exhibiting certain features of interest. Different distributions that are equally likely to generate routes that satisfy the goal(s) at hand, such as passing near your favorite café, are clustered together into a single telic state. Formally, we define telic states as classes of experience distributions that are equivalent with respect to a goal. Recall that a goal, in turn, is defined as a rank ordering of experience distributions. Telic states can therefore be formally defined as equivalence classes over experience distributions, induced by the order relation corresponding to the goal. A \emph{telic state representation} is defined as a partitioning of all possible experience distributions into goal-induced equivalence classes. As we argue below, partitioning experience distributions into telic states (i.e. goal-equivalent classes) can provide a parsimonious account of goal-directed learning, useful for addressing its behavioral, neural, and phenomenological aspects.   

\subsection{Learning with telic states}
\label{sec:learning_with_telic_states}
Within the RL framework, state representations encode the descriptive information that supports learning and decision making. It is therefore natural to ask how telic state representations support goal-directed learning and decision-making, and in what ways do they differ from more traditional reward-based approaches. To address these questions, we first introduce the notion of a policy, defined as a (generally stochastic) function mapping states to actions. Since telic states are defined as classes of experience distributions, we shall accordingly define a policy as a mapping between experience sequences and actions. In other words, for any experience or action-observation sequence, a policy determines the likelihood of each possible next action. Similarly, the environment determines, for any experience (including the agent's current action) the likelihood of each possible next observation. Since the environment is often assumed to be fixed, it follows that any policy induces a distribution over possible experiences. Thus, as telic states are classes of experience distributions, policies can be clustered into telic states as well, according to the experience distribution that they induce. Viewing both policies and telic states as classes of experience distributions allows us to quantify purposeful behavior in terms of the statistical divergence between them. In other words, under our account, goal-directed learning can be viewed as an optimization process, striving to minimize the statistical divergence between the agent's current policy and desirable telic states (for technical details, see the appendix \ref{appendix:formal_setting}). This unified perspective on policies and telic states facilitates the application of powerful statistical tools, such as information geometry \cite{ay2017information}, for developing algorithms for state representation learning in computationally-bounded agents. For example, in recent work, we leveraged the duality between goals and telic state representations to quantify the computational effort required by an agent to learn a policy that enables it to reach a desirable telic state under a given goal \cite{amir2024learning}. This allows us to cast the conjoined problems of goal selection and telic state representation in terms of a tradeoff between the granularity of the telic state representation and the computational complexity required for learning policies that can traverse all possible telic states. 

\section{Implications}
\label{sec:discussion}
\subsection{Bringing goals into view in learning}
\label{sec:other_frameworks}
A recent philosophical analysis of RL as a model of human learning argues that people's values are not merely encoded by their reward functions but are also implicitly embedded within their state representations \cite{aronowitz2024locating}. Similarly, it has recently been argued that the successes of RL urge us to recognize the fundamentally evaluative nature of the mind \cite{Haas2023-wb}. By highlighting the reciprocal relationship between goals and state representations, the telic states framework provides a formal approach that accounts for the inherent evaluative aspect of learning and representation in cognitive agents. The need for such a framework is motivated by those cases in which key common assumptions of RL, such as the identification of goals with rewards or the givenness of state representations, may not hold. As such, it extends and complements, rather than contradicts, traditional RL-based approaches to goal-directed learning in biological and artificial systems. For example, within the telic states framework, goals are expressed as preferences over complex and temporally extended statistical features of experience, with the maximization of expected accumulated scalar reward as one particular goal function. Even if it were possible to express any goal using a complex and highly tailored reward function, in many cases, this way of formalizing goals may not provide a satisfying account of why cognitive agents pursue the kinds of activities they do. Consider musicians, enjoying the intricate sensorimotor regularities within a piece of music they are playing \cite{huron2008sweet}, or basketball players enjoying the synchronization patterns between their actions and those of the other team members. Trying to explain such behaviors as completely driven by scalar reward maximization may miss something essential about the subjective experience that makes them desirable in the first place. Research in positive psychology similarly suggests that the matching between efforts (actions) and results (outcomes) leads to intrinsically desirable flow state experiences \cite{melnikoff2022computational}. The telic states framework can naturally explain why such temporally extended dependencies within sensorimotor experience streams may be intrinsically desirable. Furthermore, as we discuss below, it suggests a paradigmatic shift from an agent-centered to a goal-centered view of learning, which may shed light on phenomena such as diminished self-referential processing reported during flow and similar highly engaged motivational states \cite{csikszentmihalyi1992optimal}. 

By coupling goals and state representations, the telic states framework is complementary to several lines of active research within cognitive science and AI that aim to elucidate the computational underpinnings of preference-based and goal-conditioned reinforcement learning  \cite{furnkranz_preference-based_2012,andrychowicz2017hindsight}. For example, recent work has characterized the conditions under which optimal decision-making policies can exist with preference-based goals that cannot be expressed by a reward function \cite{carr2024conditions}. Such scenarios are common in real-world AI applications that rely on preferential feedback for sequence-to-sequence learning supervision and optimization \cite{bai2022constitutional}. Other active research domains in AI, such as transfer learning \cite{zhu2023transfer,taylor2009transfer,lu2020dynamics}, compositionality \cite{adamczyk2023bounding,adamczyk2023compositionality,jothimurugan2021compositional}, state abstraction \cite{abel2018state,pacelli2020learning,pacelli2019task} and multi-objective RL \cite{yang2019generalized,van2014multi,wray2022multi}, rely on fixed, possibly unknown, state space representations, a constraint which may be unsuitably rigid or inefficiently over-specified for flexible learning under multiple, possibly changing, goals. For example, the problem of state abstraction is naturally resolved in the telic state representation by choosing a desired resolution of representation with the tools from information theory. Optimal solutions to multi-objective RL can be found by calculating the Pareto frontier, which is a hard problem in general, while telic states can automatically incorporate multiple relevant objectives from the agent's subjective perspective. The problem of compositionality, where solutions to different problems are recomposed to solve a new problem, can be naturally formulated using the machinery of information geometry, which provides a principled way to transfer learning by interpolation between different telic states \cite{amir2024learning}. By aligning agents' preferences and experiences with their state representation in a goal-flexible way, telic states can provide a unified framework for addressing these common scenarios in current AI research. 

Endowing an agent with the capacity to learn and update state representations online during their interaction with an environment implicitly allows for (some) goal-dependence in learning. In latent cause inference, for instance, state representations are discovered by inferring the hidden probabilistic structure of the world using a Bayesian model of the relations between hidden and observed environmental variables \cite{gershman2010learning}. Assuming such inference to be goal-sensitive might allow for when and how new latent causes, or states, are inferred depending on a goal \cite{radulescu2021human}. However, while goal-sensitive hidden-state inference might cluster observations made during experience according to their inferred relevance, telic states are not discovered or inferred but rather constructed based on the current goal. While this construction process is naturally constrained by sensorimotor primitives and the causal structure of the physical world, it does not necessarily aim to uncover, or even approximate, a veridical model of the external environment. Rather, by distinguishing between goal-relevant and goal-irrelevant aspects of experience, it construes the environment in a way that is best suited for achieving its goal. Thus, unlike latent cause inference or hidden-state inference in partially observable settings, where more evidence always results in reduced uncertainty about the inferred causal structure of the world, additional observations do not necessarily decrease an agent's uncertainty about its current telic state. The environment thus indirectly shapes goals through the constraints it imposes on the space of possible experience distributions. In other words, by determining the likelihood of different observations the environment plays a dual role to the agent in structuring the space of experience distributions. An important caveat however is that the agent typically only has access to an imperfect model of the environment (similar to partial observability in the traditional RL framework) and thus the telic state representation it uses only approximates the actual space of possible experience distributions. 

Intrinsically motivated agents,recently described as \emph{autotelic}, are those that can learn to represent, generate, select, and achieve their own goals \cite{colas2022autotelic}. Constructing such agents is a major open challenge in AI. One approach has been to introduce goals as constraints over states that an agent seeks to respect, using a goal-embedding function, providing a parametrization of the state constraints owing to the goal, and a goal-achievement function, measuring the agent's progress towards it. In our framework, telic state representations encapsulate both roles without assuming a preconceived state representation structure. Instead, goals are defined directly in terms of experience preferences, which in turn underlie the formation of the (telic) state representations themselves. Cognitive scientists are also becoming increasingly interested in understanding the mechanisms underlying intrinsically motivated behavior and learning in biological agents \cite{oudeyer2007intrinsic,baldassarre2013intrinsically,karayanni2022extrinsic}. Among the proposed mechanisms are homeostatic regulation \cite{keramati2014homeostatic}, curiosity \cite{gottlieb2013information}, empowerment \cite{klyubin2005empowerment,tiomkin2024intrinsic}, predictive power \cite{ay2008predictive}, information gain \cite{vergassola2007infotaxis} and policy complexity \cite{amir2020value}. Whereas each of these frameworks highlights a particular choice of goal or motivation, the telic states framework instead takes a step back to consider how goals, generally construed, shape the way agents represent and structure their experiences.
For instance, under conditions of high cognitive load, an agent may prioritize minimizing the complexity of its actions to preserve computational resources, whereas under homeostatic stress, it might instead focus on monitoring and regulating internal parameters to maintain stability. In each of these situations, an agent would likely benefit from a different approach to structuring its experiences. Shifting the focus to goals as theoretical constructs in their own right enables us to gain insight into the representational underpinnings of purposeful behavior, separate from the algorithmic instantiation of a particular learning heuristic.

The telic states framework is related to enactivist frameworks that advocate for a nonrepresentational view of cognition as a dynamic process of enacting or ``bringing forth'' both the agent and its environment through mutual specification and codetermination within embodied and situated contexts \cite{thompson2010mind,engel2013s,varela2017embodied}. Similarly, ecological, action-oriented, and embodied views of cognition posit that mental and neural processes do not merely provide descriptive representations of states in the external world, but rather actively prescribe potential action affordances within a particular environmental and bodily context \cite{clark1998being, heft2001ecological}. More recently, these approaches have been grounded within computational frameworks such as active inference \cite{parr_active_2022} and predictive coding \cite{rao_predictive_1999}. What our approach shares with these frameworks is the view of experience as actively constructed by, and emergent from, the interaction between the agent and its environment. Telic states formally describe how this construction is inherently goal-directed or preference-based. By proposing that experiences are structured in a goal-dependent manner, telic states can serve to explain why cognition is enacted in particular ways and why generative models predict and distinguish between certain aspects of experience and not others. As a flexible account of state representation that is sensitive to an agent's goals, telic states chart a middle ground between dominant representational views and strongly non-representational enactivist views of cognition. 

What are some concrete predictions of the telic states framework and how might they differ from other RL-based approaches to state representation learning such as reward-optimized feature acquisition using deep neural networks. One class of behaviors that many standard online RL methods may struggle to account for are sequential tasks with long-range temporal dependencies and sparse rewards. For example, suppose that in order to obtain a food pellet, an agent must sequentially visit several different locations, denoted $L_1, L_2,..., L_n$ within varying time intervals $T_1, T_2,...,T_{n-1}$ between each consecutive location. While it is in principle possible to learn this task using a Markovian state representation and a suitably shaped reward function, this would generally require exponentially large state spaces, including a variable for each location indicating whether it has already been visited and a buffer for measuring each temporal interval. The sparsity of the reward signal also makes it difficult for reward-optimized  algorithms to adaptively learn the relevant task features, namely whether location $L_i$ has been visited within a time interval $T_i$ before visiting $L_{i+1}$, without some ad-hoc design of the state space. The telic states framework, on the other hand, can naturally account for the ability of the rat to determine which features need to be remembered and for how long. Assuming the rat has access to a sufficiently detailed representation of its own goal, all trajectories that pass through the required locations within the required temporal intervals are clustered as a telic state and all goal-irrelevant features are ignored. This can be implemented algorithmically using a mechanism such as prioritized experience replay \cite{schaul2015prioritized}, with the goal providing a natural criterion for selecting which experiences to prioritize. Thus, we predict that telic–based agents would learn such sequentially and sparsely rewarded tasks faster and more efficiently than traditional reward-optimized approaches. Furthermore, reward-optimized state representations may struggle to adapt to small changes in reward contingencies, such as a shift in the location of the reward \cite{zhang2018decoupling}. A telic states framework might more easily adapt to such changes, if the underlying goal remains the same. For an example of how the telic-states framework can flexibly adapt to shifting goals see appendix \ref{appendix:telic_controllability}.

\subsection{Telic states and the brain}
\label{sec:neural_basis}
What are the implications of the telic states framework for elucidating the neural underpinnings of goal-directed learning \cite{niv2016reinforcement,averbeck2022reinforcement}? Research in animals and humans has located neural signatures of state representation in the hippocampus and orbital-medial prefrontal cortex, suggesting that these brain regions play a role in the formation of cognitive maps, supporting reward learning, decision making, and planning \cite{wikenheiser2016over, wilson2014orbitofrontal, constantinescu2016organizing, behrens2018cognitive}. These cognitive maps, however, have been shown to be dynamic and goal-dependent. Indeed, contextual modulation of prefrontal neural activity implies dynamic rearrangement according to immediate task objectives, consistent with goal-dependence in neural representation in these regions across species. 
\cite{mante2013context,moneta2023task}. In the hippocampus, reward information also co-occurs with the canonical spatial encoding of this structure, which can represent abstract relational ``spaces'' tightly coupled to a reward-guided task \cite{gauthier2018dedicated, aronov2017mapping}. These hippocampal representations also display goal-sensitivity with context-dependent selective compression of navigational maps \cite{muhle2023goal}.

Telic states may also help to elucidate the role of neural simulations of sensorimotor experiences in representation learning and goal-directed behavior. For example, hippocampal place cell replay has been proposed as a potential substrate for experience-dependent simulation in goal-directed tasks \cite{foster2017replay}. Place cells,  initially thought to support the formation of invariant cognitive maps of spatial location \cite{o1978hippocampus},  were later reported also to exhibit preferential representation for goal locations \cite{hollup2001accumulation}. Furthermore, while early interpretations of place cell replay activity in the hippocampus emphasized its role in memory consolidation, later findings revealed that replay also encodes goal-oriented aspects of experience, such as the reachability between different spatial locations \cite{davidson2009hippocampal}. Similarly, entorhinal grid cells, which were initially thought to represent veridical geometric coordinates for an agent's hippocampal cognitive map, have been recently found to adapt dynamically according to its current goals \cite{boccara2019entorhinal}. Similar experience replay mechanisms have also been implicated in planning and memory preservation in human decision making, extending these principles to more complex organisms and settings \cite{wimmer2023distinct}. 

The idea that the brain is constantly generating and evaluating simulated experiences is central to current accounts of grounded cognition \cite{barsalou_simulation_2009}. Under such accounts, the brain can be viewed as a coordinated system that generates a continuous stream of multi-modal simulations, or re-enactments, of perceptual, motor, and introspective states acquired by embodied agents through interactions with their environment. These simulations provide situated conceptualizations of different experiences and are thought to occur in two phases: first, storage in long-term memory of a multi-modal activation pattern corresponding to different concepts within a particular context; and second, re-enactment, or simulation, of these patterns for prediction and representational use. The telic states framework is well-suited to be instantiated by such a mechanism of goal-oriented experience simulation. Specifically, it suggests that agents categorize their experiences into goal-oriented classes, namely telic states, thus providing a measure of proximity with respect to their goal. Planning can be similarly implemented by evoking sequences of simulated experiences and evaluating which telic states they are most likely to invoke. Taken together, these findings suggest that the brain maintains dynamic, goal-oriented representations of possible experiences, rather than merely descriptive mappings of the external environment. 

\begin{figure}[htb]
    \centering
    \includegraphics[scale=0.20]{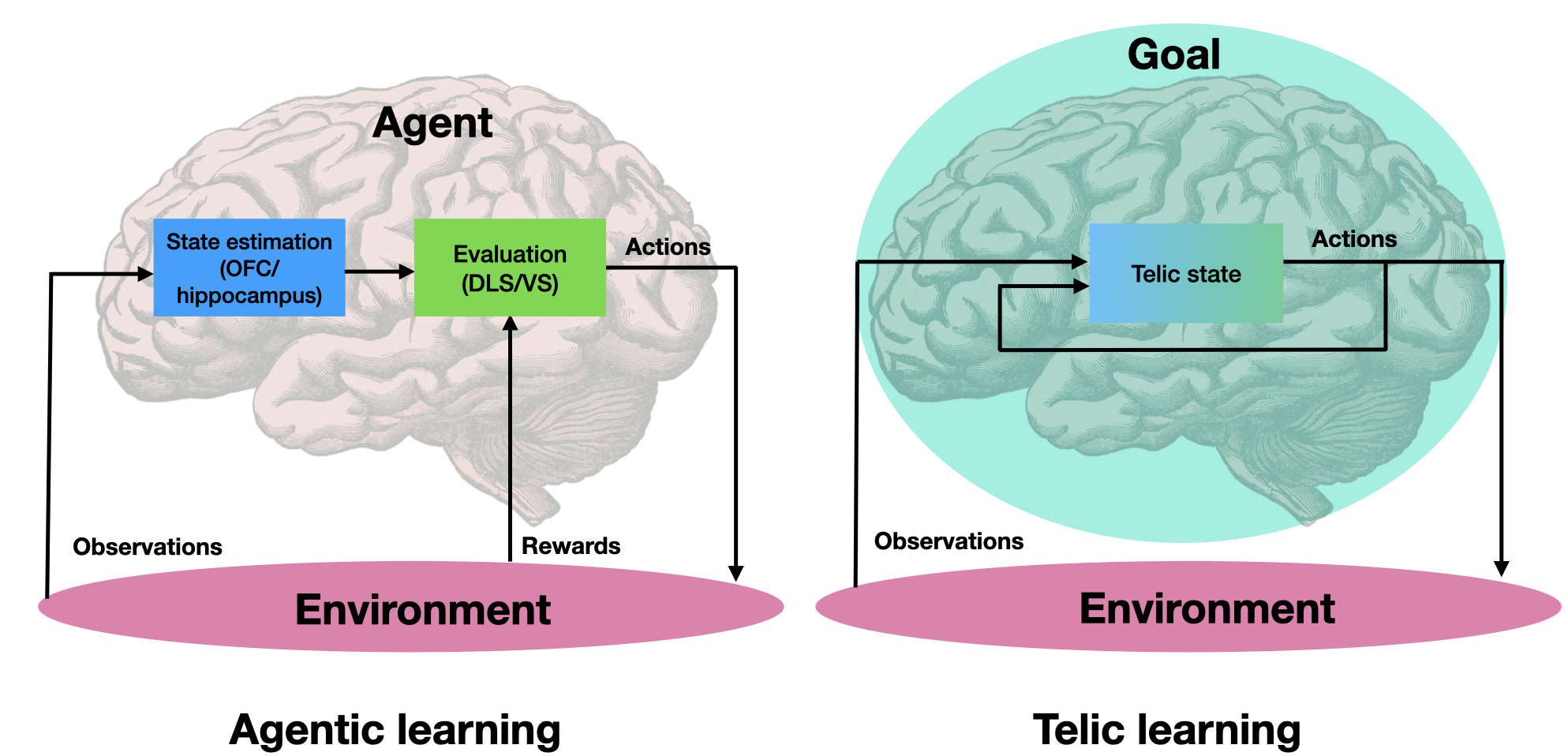}
    \caption{\textbf{Agentic vs. telic models of learning:} (Left) traditional models view agents as the fundamental learning units, with dedicated neural modules for state and value estimation.
    (Right) states are typically assumed to be estimated on the basis of the previous state and the current observation while evaluation relies on stored values for different states which are updated based on a privileged set of observations that are considered ``rewarding''.
    In the telic states framework, purposeful behavior stems from goals, which induce telic states, encapsulating descriptive and evaluative aspects of experience. Goals are represented across multiple brain regions and may even extend  beyond the brain, e.g., in cases of multiple agents pursuing a joint goal}
    \label{fig:telic_vs_agentic}
\end{figure}

We conclude this section by reviewing several concrete, albeit speculative, predictions of the telic framework regarding the neural underpinnings of state representation learning for goal-directed behavior, and how these might diverge from more traditional reward and agent-centric views (Fig.~\ref{fig:telic_vs_agentic}). The notion of a telic state suggests that any neural signature of a cognitive map would ultimately reflect an underlying intrinsic preference among different sensorimotor sequences. Thus, we predict that the brain would represent the motivational aspect of two sensorimotor sequences differently only if one is subjectively preferred over the other in some goal-relevant sense. This proposed coupling between state representation and goal-preference further suggests that individual brain regions, traditionally thought to encode either state or value related information, may in fact jointly encode a more fundamental notion of telic state. For example, we expect that the neural encoding of a task state representation in areas such as the OFC or hippocampus would always be coupled with, and modulated by, a preference related neural activity in areas such as the vmPFC or VTA. Indeed, recent findings show that value information, in prefrontal regions, is multiplexed with this map-like information, consistent with joint representation \cite{zhou2019rat,moneta2024representational}.

While we expect that experiences belonging to the same telic state would be represented differently at the sensorimotor level (for example, two different hand movements towards the same button), we predict that the neural signatures of goal-equivalent experience sequences would tend to converge towards higher, more conceptual processing levels,. In line with these predictions, a recent study suggests that action–outcome pairs are represented in the hippocampal-entorhinal system as a cognitive map organized by outcome similarity along goal-relevant dimensions, implying that functionally similar experience sequences, those that may serve similar goals, are encoded in similar ways \cite{barnaveli2025hippocampal}. 

\subsection{Where do goals come from?}
\label{sec:controllability}
Explaining how goals are formed and selected in the first place is a fundamental challenge facing any complete theory of goal-directed learning. Cognitive psychologists have identified various factors influencing goal formation and selection, such as physiological and psychological needs \cite{deci2013intrinsic},
perceived value and likelihood of success \cite{atkinson1957motivational} and cognitive resource constraints \cite{baumeister2018ego}.  
By linking goals and state representations, telic states offer a computationally principled framework for addressing the problems of goal formation and selection. Crucially, traditional value-based approaches to the problem of goal selection involve a vicious circularity. Namely, if goals determine the agent's value function, how can the latter be used to select the former? Meta-reasoning accounts attempt to address this circularity by positing a higher-order selection criterion that balances goal-dependent utility with other considerations, typically pertaining to cognitive computational resources \cite{griffiths2019doing}. Such accounts, however, face two fundamental limitations when applied to biological learning: First, they do not explain why agents employ a particular utility function and not another. Second, they do not explain why the meta-reasoning criterion takes the particular form that it does - for example, why is utility only balanced against computational, and not other forms of resources, such as metabolic, affective or social ones? Addressing these problems using meta-reasoning approaches may lead to an infinite regress: introducing meta-preferences to justify preferences would seem to then require meta-meta-preferences to justify the meta-preferences, and so on ad infinitum. However, by coupling goals and state representations the notion of telic states  may point towards a way out of this conundrum. Namely, the process of selecting a goal can be externally regulated by the properties of the telic state representation it induces. While this still requires a criterion for what makes for a telic state representation desirable, this does not necessarily lead to an infinite regress since certain properties of telic state representations are arguably desirable for any conceivable goal. In appendix \ref{appendix:telic_controllability}, we formally explore one such property by introducing the notion of telic-controllability, defined by the complexity of a policy needed to transition between all telic states, as a goal-selection criterion. This lets us recast the problem of goal selection as a tradeoff between the granularity of the resulting state representation and the computational resources required to reach all telic states. Indeed, as in meta-reasoning accounts, such as the value of information, the agent's goal, and hence its state representations, should match its capacity to utilize them. For example, during the opening stages of a game of chess, the goal of checkmating the opponent’s king would probably not yield a useful state representation since we can usually only plan a few moves ahead, making it unclear how to derive a concrete action plan from that goal. Instead, a more localized, or granular, goal, such as developing your pieces and controlling the center, would provide a more actionable, and hence useful, state representation. Rather than having to separately determine a utility (or reward) function and a criterion for balancing it with computational resource costs, the notion of telic controllability provides a unified criterion for goal selection that automatically matches the computational constraints of the agent.

Finally, formalizing the problem of goal generation using telic states may also shed new light on the phenomena of transformative experiences, defined as experiences that radically change the experiencer in both epistemic and personal ways \cite{paul2014transformative} - such as falling in love or becoming a parent. Transformative experiences, and the decision to undergo them, challenge traditional models of rational choice, since, by definition, they transform the agents that experience them, changing their preferences and goals in potentially unpredictable ways. Indeed, to the best of our knowledge, only a few attempts have been made to empirically study transformative experiences using formal rational choice models \cite{zimmerman2020models, gershman2021makes}. The notion of telic states may help  address this issue by explaining how changes in preferences also change how different choices are construed. Specifically, choosing to undergo a transformative experience may be explained in terms of the desirable properties of the ensuing telic state representation. Thus, instead of thinking of transformative experiences in terms of meta-choices over possible preferences or goals, we can view them through the lens of the telic state representation they induce. Indeed, transformative experiences can be seen as inducing a new way of  attending to the world and people who have undergone them often report seeing the world in new and unexpected ways \cite{paul2015you}.  In other words, the duality between goals and (telic) state representations may offer a way of circumventing the problem of infinite regress mentioned above, arising when trying to provide a rational basis for transformative decisions. However, viewing goals and state representations as two sides of the same coin can provide a rational criterion for choosing goals by considering the properties of the state representation they induce. For example, an agent with limited computational resources would prefer a simple telic representation (one that can be utilized by simple policies).

\section{Conclusion: from agentic to telic models of learning}

The telic states framework invites us to reexamine some of our basic assumptions regarding the origin, structure, and role of goals in shaping the phenomenology of purposeful behavior. It offers a perspective shift, from a view centered on reward-driven agents as enduring and centralized loci of control (Fig. \ref{fig:telic_vs_agentic}, left), to one where subjective experience streams are dynamically shaped by goal-dependent representational structures, or telic states (Fig. \ref{fig:telic_vs_agentic}, right). 

Interestingly, recent conceptual and empirical advances in biology and cognitive science have called into question the fecundity of the notion of a centralized unit of of control, namely an agent (=a ``Self''), for explaining the phenomenology of purposeful behavior across diverse substrates and spatiotemporal scales  \cite{levin2022technological,difrisco2024biological}. Biological learning occurs over multiple scales, from the cellular level \cite{gershman2021reconsidering} to that of whole ecosystems \cite{power2015can}, with corresponding goals and representational structures operating on each level. A notion of agency grounded on dynamically extended goals, rather than a fixed agent, may thus be better suited to capture the diversity of possible intelligences and explain their origin. By shifting the focus from agents to goals, our framework provides a unified foundation for studying diverse forms of intelligence, both natural and artificial. The telic states framework provides a step towards a formal framework of goal-directed learning, grounded in the statistical structure of subjectively structured experience streams, without invoking the notion of fixed and enduring agents. Ultimately, we hope that our framework will help pave the way towards more complete views on cognition and purposeful behavior, ones that can reconcile ancient philosophical and modern computational views on agency and experience. 

\appendix
\section{Formal setting}
\label{appendix:formal_setting}

\subsection{Telic states as goal-equivalent experiences}
We assume the setting of a perception-action cycle, or streams of observation-action pairs representing the interaction between the agent and its environment. We denote by $\mathcal{O}$ and $\mathcal{A}$ the set of possible observations and actions, respectively. An experience sequence, or \emph{experience} for short, is a finite sequence of observation-action pairs: $h=o_1,a_1,o_2,a_2,...,o_n,a_n$. For every non-negative integer, $n\geq0$, we denote by $\mathcal{H}_n\equiv(\mathcal{O}\times\mathcal{A})^n$ the set of all experiences of length $n$. The collection of all finite experiences is denoted by $\mathcal{H}=\cup_{n=1}^{\infty}\mathcal{H}_n$. In non-deterministic settings, it will be useful to consider distributions over experiences rather than individual experiences themselves and we denote the set of all probability distributions over finite experiences by $\Delta(\mathcal{H})$. 
Following \cite{bowling2022settling}, we define a \emph{goal} as a binary preference relation over experience distributions. For any pair of experience distributions, $A,B\in\Delta(\mathcal{H})$, we write $A\succeq_gB$ to indicate that experience distribution $A$ is weakly preferred by the agent over $B$ (i.e., that $A$ is at least as desirable as $B$) with respect to goal $g$. 
When $A\succeq_gB$ and $B\succeq_gA$ both hold, $A$ and $B$ are equally preferred with respect to $g$, denoted as $A\sim_gB$. We observe that $\sim_g$ 
is an equivalence relation, i.e., it satisfies the following properties:
\begin{itemize}
    \item Reflexivity: $A\sim_gA$ for all $A\in\Delta(\mathcal{H}).$
    \item Symmetry: $A\sim_gB$ implies $B\sim_gA$ for all $A,B\in\Delta(\mathcal{H}).$
    \item Transitivity: if $A\sim_gB$ and $B\sim_g C$ then $A\sim_g C$ for all $A,B,C\in\Delta(\mathcal{H}).$
\end{itemize}
Therefore, every goal induces a partition of $\Delta(\mathcal{H})$ into disjoint sets of equally desirable experience distributions. For goal $g$, we define the goal-directed, or \emph{telic}, state representation, $\mathcal{S}_g$, as the partition of experience distributions into equivalence classes it induces:
\begin{equation}
    \mathcal{S}_g= \Delta(\mathcal{H})/\sim_g.
\end{equation}
In other words, each telic state represents a generalization over all equally desirable experience distributions. This definition captures the intuition that agents need not distinguish between experiences that are equivalent (in a statistical sense) with respect to their goal. Furthermore, since different telic states are, by definition, non-equivalent with respect to $\succeq_g$, the goal $g$ also determines whether a transition between any two telic states brings the agent in closer alignment to, or further away from its goal.
\subsection{Learning with telic states}
How can telic state representations guide goal-directed behavior? To address this question, we start by defining a \emph{policy}, $\pi$, as a distribution over actions given the past experience sequence and current observation:
\begin{equation}
\label{eq:policy_def_orig}
\pi(a_i|o_1,a_1,...,o_i).
\end{equation}
Analogously, we define an \emph{environment}, $e$, as a distribution over observations given the past experience sequence: 
\begin{equation}
e(o_i|o_1,a_1,...,a_{i-1}).
\end{equation}
The distribution over experience sequences can be factored, using the chain rule, as follows: 
\begin{equation}
     P_\pi(o_1,a_1,...,o_n,a_n)=P(o_1,a_1,...,o_n,a_n|e,\pi)=\\
    \prod_{i=1}^n e(o_i|o_1,a_1,...,a_{i-1})\pi(a_i|o_1,a_1,...,o_i).
\end{equation} 
Typically, the environment is assumed to be fixed, and hence not explicitly parameterized in $P_\pi(h)$ above.
Our definition of telic states as goal-induced equivalence classes can now be extended to equivalence between policy-induced experience distributions as follows: 
\begin{equation}
    \pi_1\sim_g\pi_2\iff P_{\pi_1}\sim_gP_{\pi_2}.
\end{equation}
The question we are interested in can now be stated as follows: how can an agent learn an efficient policy for reaching a desired telic state? In other words, how can the agent's policy be updated to increase its likelihood of generating experiences that belong to a certain desirable telic state, $S_i\in\mathcal{S}_g$? 
To answer this question, we begin by writing down the empirical distribution of $N$ experience sequences generated by policy $\pi$:
\begin{equation}
    \hat{P}_\pi(h)=\frac{|\{k:h_k=h\}|}{N}.
\end{equation}
We would like to estimate the probability that $ \hat{P}_\pi(h)$ belongs to telic state $S_i$, and update $\pi$ to increase this probability. 
A fundamental result from large deviation theory, known as Sanov's theorem \cite{cover1999elements}, shows that this probability decays exponentially with a rate of 
\begin{equation}
\label{eq:telic_distance}
R=\min_{P\in S_i}D_{KL}(P||P_\pi).
\end{equation}
Since $R$ determines the probability that experiences sampled from $P_\pi$ belong to telic state $S_i$, we refer to it as the \emph{telic distance} from $\pi$ to $S_i$. Assuming now the agent's policy can be expressed using some parameterization $\theta$, the following policy gradient method updates $\pi_\theta$ in a way that minimizes the telic distance, i.e., maximizes the likelihood of generating experiences belonging to telic state $S_i$:
\begin{equation}
    \label{eq:policy_grad_step}
    \theta_{t+1}=\theta_t-\eta\nabla_\theta D_{KL}(P_i^\star||P_{\pi_\theta}),
\end{equation}
where $\eta>0$ is a learning rate parameter and,
\begin{equation}
\label{eq:info-proj}
P_i^\star = arg\min_{P\in S_i} D_{KL}(P||P_\pi),
\end{equation} is called \emph{information projection} of $P_\pi$ onto $S_i$, i.e., the distribution in $S_i$ which is closest, in the KL sense, to $P_\pi$. Equation \ref{eq:policy_grad_step} thus describes a general policy gradient method for learning with telic state representations. 

\subsection{Illustrative example: the two-armed bandit}
To illustrate our proposed learning algorithm, we compute the goal-directed policy gradient for a fully-tractable bandit learning problem and show that, in this simple case, minimizing telic distance yields a commonly reported empirical choice strategy known as probability-matching. We consider a two-armed bandit in which the set of actions is defined as of choosing a left ($L$) or right ($R$) lever and the observations are winning ($1$) or losing ($0$): 
\begin{equation}
    \mathcal{A}=\{L,R\}, \ \mathcal{O}=\{1,0\}.
\end{equation}
For simplicity we consider a past-independent policy, $\pi_\theta$, that is parameterized by the probability of choosing action $L$:
\begin{equation}
    \pi_\theta(L)=\theta,\ \pi_\theta(R)=1-\theta.
\end{equation}
The environment $e$ is specified by the probabilities of winning when choosing $L$ or $R$, denoted $p_L$ and $p_R$, respectively:
\begin{equation}
    e(1|L)=p_L, \ e(0|L)=1-p_L; \ e(1|R)=p_R, \ e(0|R)=1-p_R.
\end{equation}
The likelihood that an experience sequence, $h$, will be generated by the policy induced distribution $P_{\pi_\theta}$ can be expressed as:
\begin{equation}
\label{eq:seq_prob}
    P_{\pi_\theta}(h)=\theta^{N^h_L}(1-\theta)^{N^h_R}p_L^{N^h_{L,1}}(1-p_L)^{N^h_L-N^h_{L,1}}p_R^{N^h_{R,1}}(1-p_R)^{N^h_R-N^h_{R,1}},
\end{equation}
where $N^h_L,N^h_R$ are the number of times the agents selected the $L$ and $R$ actions, respectively, and $N^h_{L,1},N^h_{R,1}$ are the number of ``win'' observations following $L$ and $R$ choices, respectively. For simplicity, we assume that the agents goal is to reach a specific number of wins, so that two policies are equivalent if and only if the expected number of wins obtained by following both is equal:
\begin{equation}
\pi_{\theta_1}\sim_g \pi_{\theta_2} \iff \mathbb{E}_{P_{\pi_{\theta_1}(h)}}\left(\sum_i^{N}\mathbf{1}_{h_i=1}\right)= \mathbb{E}_{P_{\pi_{\theta_2}}(h)}\left(\sum_i^{N}\mathbf{1}_{h_i=1}\right),
\end{equation}
where $N=N^h_L+N^h_R$ is the total number of action/observation pairs and $\mathbf{1}_{h_i=1}$ denotes an indicator function which is one if the $i$th observation in $h$ is $1$ and zero otherwise. Thus, for every $j=1,...,N$, the telic state $S_j$ is defined simply as the set of all experience distributions with an expected number of wins equal to $j$:
\begin{equation}
    S_j=\{P(h):\text{ s.t. }\mathbb{E}_{h\sim P}\left(\sum_i^{N}\mathbf{1}_{h_i=1}\right)=j\}.
\end{equation}
The telic distance (Eq.\ref{eq:telic_distance}) between a policy ${\pi_\theta}$ and $S_j$ is given by:
\begin{equation}
D_{KL}(S_j||P_{\pi_\theta})=\min_{P\in S_j} D_{KL}(P||P_{\pi_\theta})=\sum_hP^\star_j(h)\log\frac{P^\star_j(h)}{P_{\pi_\theta}(h)},
\end{equation}
where $P^\star_j(h)$ is the distribution in $S_j$ closest to $P_\pi(h)$ in the KL sense, as defined in Eq.~\ref{eq:info-proj} above.
Using Eq.~\ref{eq:seq_prob} we can compute the telic distance gradient:
\begin{equation}
    \nabla_\theta D_{KL}(S_i||P_{\pi_\theta})=\frac{\sum_hP^\star_j(h)N_R^h}{1-\theta}-\frac{\sum_hP^\star_j(h)N_L^h}{\theta},
\end{equation}
so that the optimal policy for reaching $j$ wins, $\pi_{\theta^*_j}$, is given by:
\begin{equation}
    \theta_j^*=\mathbb{E}_{h\sim P^\star_j}(\frac{N^h_L}{N^h_L+N^h_R}).
\end{equation}
In words, the policy maximizing the likelihood of reaching telic state $S_j$, is one matching the expected choice probability of $P^{\star}_j(h)$. Interestingly, a similar ``probability-matching''  strategy was found in human iterated binary choice behavior \cite{erev2005adaptation}.

\subsection{Closing the loop: telic state conditioned policies}
Above we have assumed that the policy depends on the full past experience but this assumption can be relaxed. Within the current framework, we assume that the agent maintains an estimate of the most likely telic state it is currently in and updates it at each time point. Concretely, given a goal $g$ and a past experience sequence at time $t$, $h_t=o_1,a_1,...,o_t$, the agent can estimate its current telic state, i.e., the equivalence class of the experience distribution most likely to have generated $h_t$:
\begin{equation}
    \label{eq:telic_state_estimation}
    \hat{S}_t(h_t)=[\arg\max_{P\in\Delta(  \mathcal{H}_t)} P(h_t)]_{\sim_g}.
\end{equation}
The policy can now be expressed in terms of the estimated telic state:
\begin{equation}
\label{eq:policy_def_telic}
\pi(a_t|\hat{S}_t(h)),
\end{equation}
so that in choosing actions the agent generalizes over past experiences that are estimated to originate from the same telic state. Since the borders between telic states are determined by the goal, the same experience may be assigned to different telic states under different goals. This clustering of past experience into estimated telic states is lossy: it ignores goal-irrelevant information, and is not necessarily Markovian. Thus, while it may not be optimal in the Bayesian sense, it provides a self contained account of how goals, i.e., preferences over experience distributions, generate intrinsic (telic) state representation, which in turn provide a foundation for action selection and learning. 

\subsection{Experience features and discrimination sensitivity}
\label{sec:sensitivity}
A natural way of representing goals, i.e., preferences over experience distributions, is by comparing the likelihood that experiences generated from different distributions will belong to some subset $\Phi_g\subset\mathcal{H}$ representing some desired property of experiences. For example, for the goal of solving a maze, $\Phi_g$ might be the set of all experiences, i.e., path trajectories, that reach the exit. Formally, for two experience distributions, $A$ and $B$, the agent will prefer the one that is more likely to generate
experiences belonging to $\Phi_g$:
\begin{equation*}
    \label{eq:pref_by_property}
    A\succeq_gB : \sum_{h\in\Phi_g}A(h)\geq \sum_{h\in\Phi_g}B(h).
\end{equation*}
The sensitivity parameter, $\epsilon$, effectively determining the maximum difference, in terms of desirable outcome likelihoods, that the agent is willing to ignore in order to reduce representational complexity. In the maze example, experience distribution $A$ would be preferred over $B$ if it is more likely to generate trajectories that reach the exit. 
Importantly, Eq.~\ref{eq:pref_by_property} implies that $A$ and $B$ are equivalent only when $\sum_{h\in\Phi_g}A(h)$ and $\sum_{h\in\Phi_g}B(h)$ are precisely equal, which is unlikely in realistic, noisy environments. A more reasonable assumption is that agents can discriminate sampling likelihoods at some finite sensitivity level, $\epsilon>0$, such that: 
\begin{equation}
\label{eq:equiv_granularity}
A\sim^{(\epsilon)}_g B \iff |\sum_{h\in\Phi_g}A(h)-\sum_{h\in\Phi_g}B(h)|\leq\epsilon. 
\end{equation}
In the maze example, this means that two trajectory distributions are considered equivalent if their respective likelihoods of generating exit-reaching trajectories are within $\epsilon$ of each other.   
As we shall see in the following sections, the discrimination sensitivity parameter, $\epsilon$, controls the tradeoff between the granularity of a telic state representation and the policy complexity needed to reach all telic states. 

\section{Telic-controllability and the goal selection problem}
\label{appendix:telic_controllability}
In this section, we introduce the notion of \emph{telic-controllability}, a joint property of an agent and a telic state representation, that characterizes whether or not the agent is able to reach all possible telic states using complexity-limited policy update steps. Towards this, we first define an agent's \emph{policy}, $\pi$, as a distribution over actions given the past experience sequence and current observation: $\pi(a_i|o_1,a_1,...,o_i)$.
Assuming a fixed environment, the definition of telic states as goal-induced equivalence classes induces corresponding equivalence classes of policy-induced experience distributions as follows: 
\begin{equation}
    \label{eq:policy_equivalence}
    \pi_1\sim_g\pi_2\iff P_{\pi_1}\sim_gP_{\pi_2}.
\end{equation}
As detailed above, this mapping between policies and telic states provides a unified account of goal-directed learning in terms of the statistical distance between policy-induced distributions and desired telic states. To explore this notion, we introduce a new property -- telic-controllability -- that plays a central role in the following sections. A representation is called telic-controllable if any state can be reached using a finite number, $N$, of complexity-limited policy updates, starting from the agent's default policy, $\pi_0$, where the complexity of a policy update step is quantified by the Kullback-Leibler (KL) divergence between the post and pre-update step policies. Formally, we have the following:
\begin{definition*}[telic-controllability]
A telic-state representation, $\mathcal{S}_g$, induced by the goal, $g$, is \emph{telic-controllable} with respect to a default policy, $\pi_0$, and a policy complexity capacity, $\delta\ge0$, if the following holds:
\begin{equation}
\label{eq:telic_controllability}
\begin{split}
&\forall S\in \mathcal{S}_g\; \exists \{\pi_t,S_t\}_{t=0}^N,N>0  \text{ s.t. } \forall t<N\\
&\big(S_t=[P_{\pi_t}]_{\sim_g}\big) \wedge \big(D_{KL}(P_{\pi_{t+1}}||P_{\pi_{t}})\leq \delta\big)\wedge \big(S_N=S\big),
\end{split}
\end{equation}
\end{definition*}
where $[P_{\pi_t}]_{\sim_g}$ is the goal-induced equivalence class, i.e., telic state, containing $P_{\pi_t}$. This definition generalizes the familiar control theoretic notion of controllability in two important ways. First, it applies to telic states, i.e., classes of distributions over action-outcome trajectories, rather than by n-dimensional vectors -- the standard control theoretic setting. Second, it takes into account the complexity capacity limitations of the agent, using information theoretic quantifiers to constrain the maximal complexity of policy update steps an agent can take in attempting to reach one telic state from another. As illustrated in the next section, telic-controllability is a desirable property since it means that agents can flexibly adjust to shifting goals using bounded policy complexity resources.
 
 \subsection{State representation learning algorithm}
 \label{sec:learning_algorithm}
 A central feature of our approach is the duality it establishes between goals and state representations. In this section, we utilize this duality to develop an algorithm for learning a telic-controllable state representation, or, equivalently, finding a goal that produces such a state representation. The algorithm receives as inputs the agent's current goal, $g$ (represented, e.g., by an ordered set of desired experience features), and default policy, $\pi_0$, along with its policy complexity capacity, $\delta$, and the discrimination sensitivity parameter $\epsilon$. Its output consists of a new goal $g'$ such that $\mathcal{S}_{g'}$ is telic controllable with respect to $\pi_0$ and $\delta$. 
 The main idea is to split any unreachable telic state, $S$, i.e., one that cannot be reached from $\pi_0$ using policy update steps with complexity less than $\delta$. State splitting is accomplished by generating a new, intermediate, telic state, $S_M$, lying between the agent's default policy induced distribution, $P_{\pi_0}$, and its information projection on the unreachable telic state, i.e., the distribution $P^*\in S$ that is closest to $P_{\pi_0}$, in the KL sense. The intermediate telic state, $S_M$, is then defined as the set of all distributions that are $\epsilon$-equivalent to $P_M$ (Eq.~\ref{eq:equiv_granularity}), where $P_M$ is the convex combination of $P^*$ and $P_{\pi_0}$ lying at a KL distance of $\delta$ from $P_{\pi_0}$.  After generating the new state, $S_M$, the goal is updated to reflect the proper ordering between the default policy state $S_0$, the intermediate state $S_M$, and the originally unreachable state $S$, such that elements of $S_M$ are between $S_0$ and $S$ in terms of preference. Pseudocode for the learning algorithm is provided in Algorithm~\ref{alg:learn_telic_sr}. The algorithm makes use of an auxiliary procedure, \textproc{FindReachableStates} (Algorithm~\ref{alg:get_reachable_states}), to find all reachable states, given the agent's goal, $g$, default policy, $\pi_0$, and policy complexity constraint, $\delta$. This auxiliary procedure performs a recursive search, similar to depth-first search methods, attempting to find policies that are closest, in the KL sense, to currently unreachable telic states, while still sufficiently close to the agent's current policy, as not to exceed the policy complexity capacity. Its main optimization step (line 3) can be implemented, e.g., using policy gradient over the information projection of $P_{\pi_0}$ on $S$. 
\begin{algorithm}[ht]
\caption{Telic-controllable state representation learning}\label{alg:learn_telic_sr}
\begin{algorithmic}[1]
\Require{$\pi_0 \text{: default policy, } g \text{: current goal, } \newline \delta \text{: policy complexity capacity, } \epsilon \text{: sensitivity.}$}
\Ensure $g' \text{: new goal such that } \mathcal{S}_{g'}$ is telic-controllable with respect to $\pi_0$ and $\delta$
\State $\mathcal{R} \gets [P_{\pi_0}]_{\sim_{g}}$ \Comment{initialize reachable state set}
\State $g'\gets g$  \Comment{initialize new goal}
\While {$\mathcal{R} \ne \mathcal{S}_{g'}$}
    \State $\mathcal{R} \gets \Call{FindReachableStates}{\pi_0,g',\delta}$ \Comment{see algorithm~\ref{alg:get_reachable_states} below}
    \For{$S\in \mathcal{S}_{g'}\setminus\mathcal{R}$} \Comment{for each unreachable state}
        \State $P^{*} \gets \arg\min_{P\in S} D_{KL}(P||P_{\pi_{0}})$ \Comment{information projection of $P_{\pi_0}$ on $S$}

       \State $M = \arg\max_{t\in[0,1]} t \text{ s.t. } D_{KL}\big((t P^{*}+(1-t)P_{\pi_{0}})||P_{\pi_0}\big)\leq\delta$ 
        
        \State $P_M = M P^{*}+(1-M)P_{\pi_{0}}$ \Comment{convex combination of $P^*$ and $P_{\pi_0}$}
        \State $S_M \gets \{P : P\sim^{(\epsilon)}_g P_M\}$ \Comment{$\epsilon$-neighborhood of $P_M$}
        \If{$P_{\pi_{0}}\leq_gP^{*}$} \Comment{update goal with preference order for $S_M$}
            \State $g'\gets g' \cup \{(p,q)_{\leq_{g'}}\in S_M\times S \}\cup \{(r,p)_{\leq_{g'}}\in S_0\times S_M\}$
        \ElsIf{$P^{*}\leq_gP_{\pi_{0}}$}
            \State $g'\gets g' \cup \{(q,p)_{\leq_{g'}}\in S\times S_M \}\cup \{(p,r)_{\leq_{g'}}\in S_M\times S_0 \}$
        \EndIf
    \EndFor
\EndWhile
\State \Return{$g'$}
\end{algorithmic}
\end{algorithm}

\begin{algorithm}
\caption{Finding reachable states }\label{alg:get_reachable_states}
\begin{algorithmic}[1]
\Require{$\pi_0 \text{: initial policy, } g \text{: goal, } \newline \delta \text{: policy complexity constraint.}$}
\Ensure{all telic states in $\mathcal{S}_g$ reachable from ${\pi_0}$ by $\delta$-complexity limited policy update steps}
\Procedure{RecursiveReach}{$\pi,g,\delta,\mathcal{R}$}
    \For{$S\in \mathcal{S}_g\setminus\mathcal{R}$} \Comment{for every unreached state $S$}
        \State $\pi_\theta\gets\arg\min_\theta D_{KL}(S||P_{\pi_{\theta}}) \text{ s.t. } D_{KL}(P_{\pi_{\theta}}||P_{\pi})\leq\delta$ \Comment{optimize policy to reach $S$}
        \If{$[P_{\pi_\theta}]_{\sim_g}\notin\mathcal{R}$} \Comment{if new state reached}
         \State $\mathcal{R}\gets \mathcal{R}\cup[P_{\pi}]_{\sim_g}$ \Comment{add current state to reachable set}
        \State $\mathcal{R} \gets \Call{RecursiveReach} {\pi_\theta,g,\delta,\mathcal{R}}$  \Comment{continue from current state}
        \EndIf
    \EndFor
    \State \Return{$\mathcal{R}$}
\EndProcedure
\Procedure{FindReachableStates}{$\pi_0, g, \delta$}
    \State $\mathcal{R}_0 \gets [P_{\pi_0}]_{\sim_g}$ \Comment{initialize reachable set} 
    \State $\mathcal{R} \gets$ \Call{RecursiveReach} {$\pi_0,g,\delta,\mathcal{R}_0$} \Comment{try to reach all states recursively}
    \State \Return{$\mathcal{R}$} \Comment{return set of reachable states}
\EndProcedure

\end{algorithmic}
\end{algorithm}

\subsection{Illustrative example: dual goal navigation task}
\label{sec:example}
\begin{figure}[ht]
    \centering
    \includegraphics[width=0.75\textwidth]{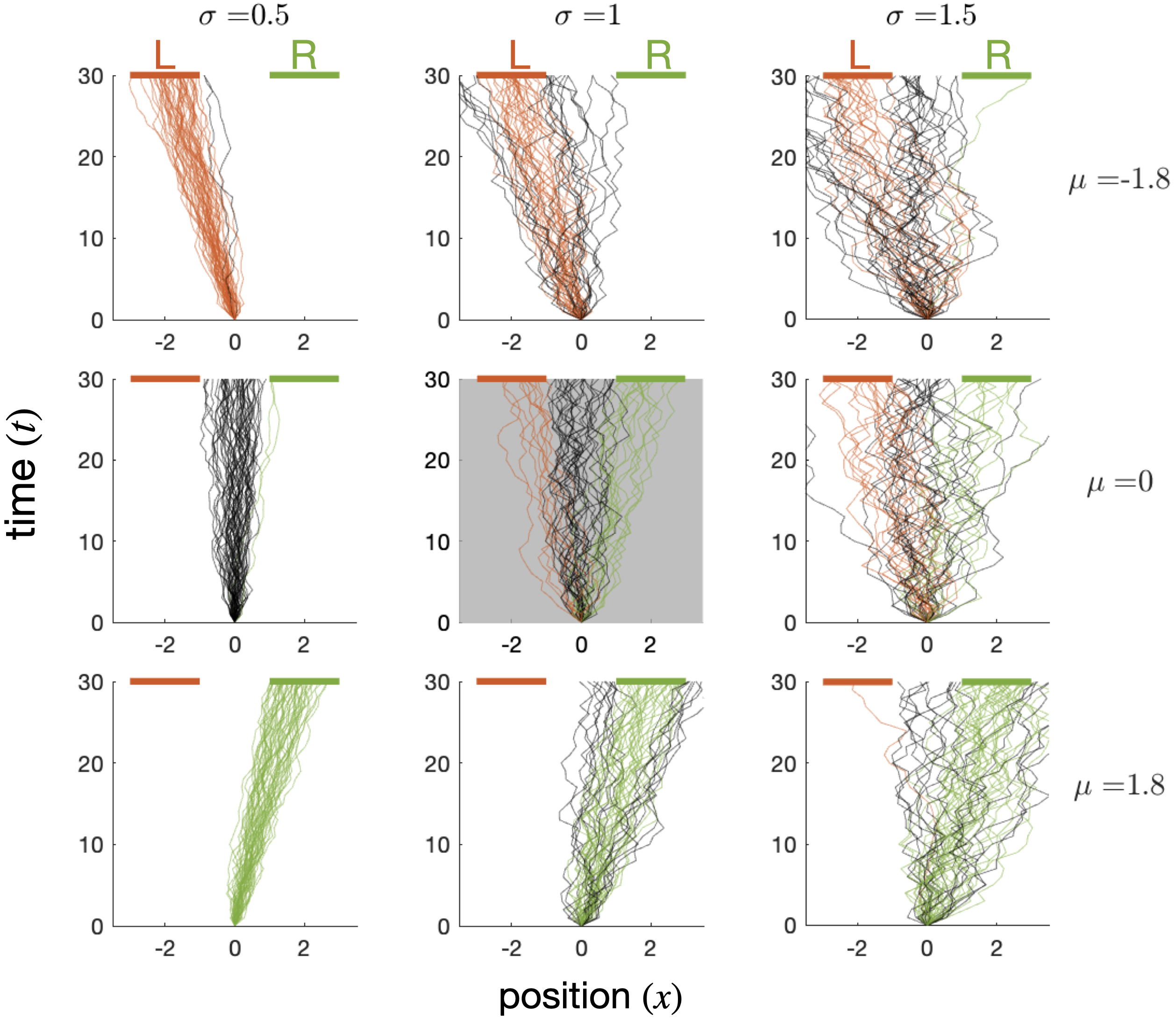}
    \caption{\small \textbf{Dual goal navigation task:} each tile shows 500 one-dimensional random walk trajectories of length $T=30$, generated by a Gaussian policy parameterized by the mean ($\mu$) and standard deviation ($\sigma$) of position update step (x-axis) across time (y-axis). Regions of interest $R$ and $L$ consist of line segments centered around $x_R=2$ and $x_L=-2$, shown as green and red lines respectively at $T=30$. Trajectories reaching one of the goals are plotted in the corresponding color, illustrating the relationship between policy parameters and goal reaching likelihoods. The default policy, $(\mu_0,\sigma_0)=(0,1)$, shown in the center gray tile, is equally likely to reach $R$ and $L$.}
    \label{fig:random_walk_traj}
\end{figure}

In this section, we illustrate the proposed telic state representation framework and learning algorithm outlined above using a simple navigation task in which an agent performs a one dimensional random walk, starting at location $x_0=0$, with the goal of reaching one of two non overlapping regions of interest after a fixed number, $T=30$, of steps. 
The agent's policy is defined as a stochastic mapping between its current and next position and is parameterized by the mean and standard deviation ($\mu$ and $\sigma$, respectively) of a Gaussian update step: $\pi(x_{t+1}|x_t;\mu,\sigma)=x_t+\eta_t,\quad \eta_t\sim\mathcal{N}(\mu,\sigma).$
For brevity, we denote by $\pi(\mu,\sigma)$  a policy with a $\mathcal{N}(\mu,\sigma)$ distributed noise term. A graphical illustration of the task and sample trajectories for different policies is shown in Fig.~\ref{fig:random_walk_traj}.
\par
\begin{figure}[ht]
        \centering
            \includegraphics[width=0.75\textwidth]{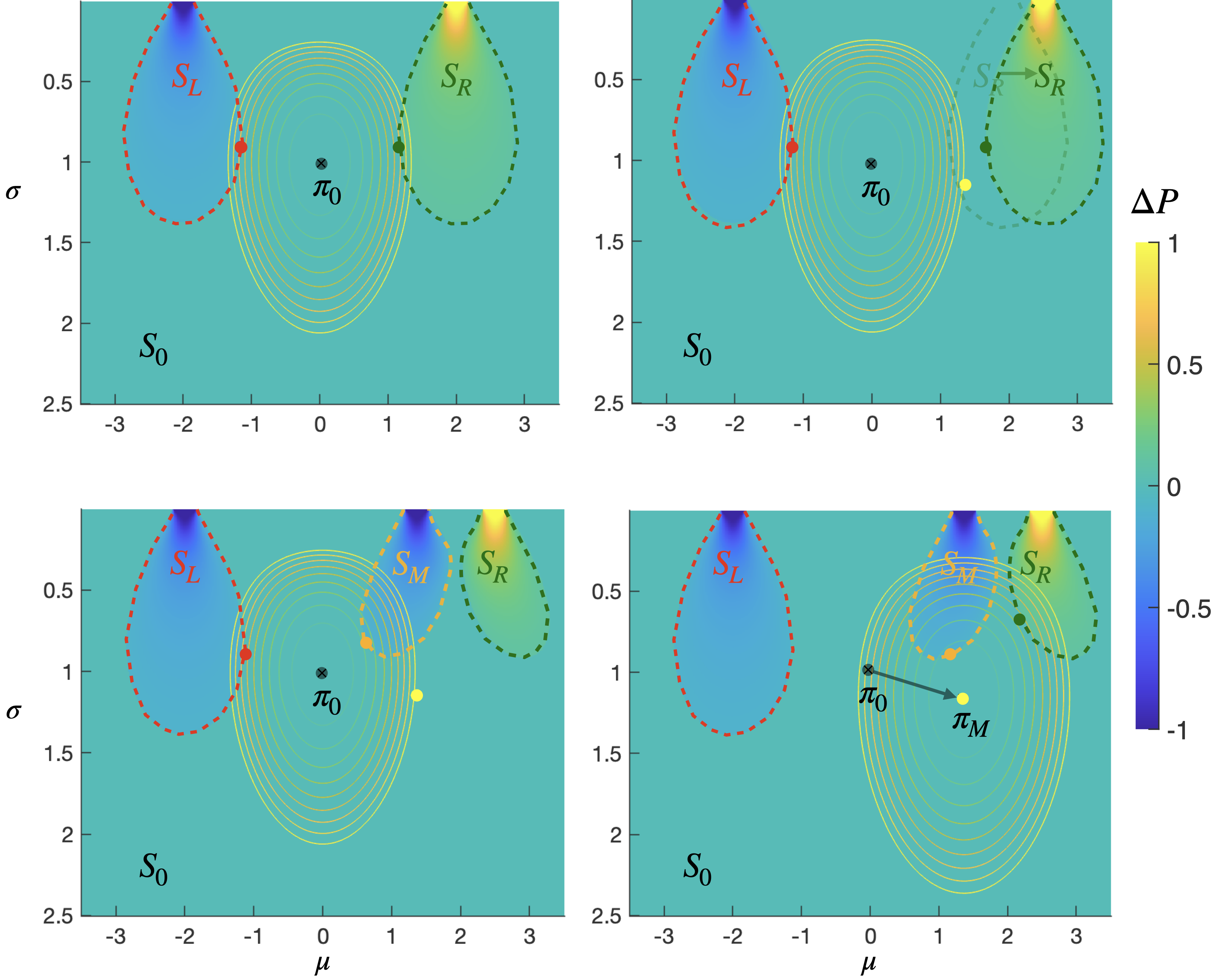} 
        \caption
        {\small \textbf{Telic state representation learning for navigation task with shifting goals:} points in $(\mu,\sigma)$ policy space colored by the difference between their probability of reaching unit length regions, $R$ and $L$, centered around $2$ and $-2$ respectively, at time $T=30$. \textbf{Top left:} telic states $S_L$ and $S_R$ (outlined by red and green dashed lines, respectively) consist of policies that are more likely to reach the corresponding region by a threshold of $\epsilon=0.1$ or more. Contour lines indicate isometric policy complexity levels, relative to the default policy $\pi_0:(\mu_0=0,\sigma_0=1)$ (black dot), for a capacity bound of $\delta=1$ bit. Green and red dots show the information projection of $\pi_0$ on $S_R$ and $S_L$ respectively, i.e., the policies each telic state closest to $\pi_0$ in KL-divergence~\textbf{Top right:} shifting the center of $R$ to $2.5$, renders $S_R$ unreachable from $\pi_0$ with $\delta$ bounded policy complexity. The policy $\pi_M:(\mu_M,\sigma_M)$ (yellow dot) is the one closest to $S_R$ while still within the complexity capacity of the agent.~\textbf{Bottom left:} splitting $S_R$ by inserting an intermediate telic-state, $S_M$, centered around $\mu_M$. By construction, the nearest distribution to $\pi_0$ in $S_M$, in the KL sense (orange dot), is within the agent's complexity capacity. ~\textbf{Bottom right:} both $S_M$ and $S_R$ are reachable with respect to the agent's new default policy, $\pi_M(\mu_M=1.37,\sigma_M=1.15)$ (see algorithm~\ref{alg:learn_telic_sr} for details); the new telic state representation $\{S_0,S_L,S_M,S_R\}$ is telic controllable with respect to $\pi_0(0,1),\,\delta=1,$ and $N=1$.} 
    \label{fig:phase_plots}
    \end{figure}

Since the sum of normally distributed variables is also normally distributed, a policy $\pi(\mu,\sigma)$ induces a Gaussian distribution over the final location of the agent:
\begin{equation}
\label{eq:random_walk_final_loc_dist}
p(x_T\mid x_0=0;\mu,\sigma)=\mathcal{N}(T\mu,\sqrt{T}\sigma).
\end{equation}
To account for goal-directed behavior, we define a right and a left region of interest, $R$ and $L$, consisting of unit radius segments centered around $x_R=2$ and $x_L=-2$ respectively. Thus, $R=[R_1,R_2]=[1,3]$ and $L=[L_1,L_2]=[-3,-1]$. For the purpose of this example, we assume that the agent wants to reach $R$ but avoid $L$, at time $T$. For example, for a rodent navigating a narrow corridor, $R$ and $L$ may indicate segments of the corridor where a reward (e.g., food) and a punishment (e.g., air puff) are administered, respectively.
We can express the agent's goal in terms of preferences over policies by defining $\Delta P(\mu,\sigma)=p(x_T\in R\mid \mu,\sigma)-p(x_T\in L\mid \mu,\sigma)$ as the difference between the probabilities that the agent will reach regions $R$ and $L$ at time $T$, with a policy $\pi(\mu,\sigma)$. The agent's goal can now be defined as a preference for policies with higher $\Delta P$ values. However, as explained above, due to the agent's finite discrimination resolution, it can only detect whether $\Delta P$ is above or below the sensitivity threshold, $\epsilon$. Thus, using Eq.~\ref{eq:policy_equivalence}, the agent's goal, $g$, can be expressed by the following preference relation over policies, where we denote, for brevity,  $\pi(\mu_i,\sigma_i)$ and $\Delta P(\mu_i,\sigma_i)$ as $\pi_i$ and $\Delta P_i$, respectively, for $i=1,2$:
\begin{equation}
\label{eq:pref_structure_R}
\begin{split} 
&\pi_1\succeq_g \pi_2\iff\big(\Delta P_1\geq\epsilon\geq\Delta P_2\big)\lor
\big(\Delta P_1\geq-\epsilon\geq\Delta P_2\big),
\end{split}
\end{equation}
where first term on the r.h.s. of Eq.~\ref{eq:pref_structure_R} captures the \emph{desirability} of $R$ -- the agent prefers policies that have a probability \emph{higher} than $\epsilon$ of reaching $R$ over ones that do not; while the second term captures the \emph{undesirability} of $L$ -- the agent prefers policies that have a probability \emph{lower} than $\epsilon$ to reach $L$ than ones that do not. We recall that telic states can be defined by policies that are similarly preferred, under the agent's discrimination threshold, $\epsilon$, which determines the borders between the resulting telic states. The telic state representation for the goal $g$ defined by Eq.~\ref{eq:pref_structure_R}, and a threshold parameter of $\epsilon=0.1$ is visualized in Fig.~\ref{fig:phase_plots} (top left). Telic state $S_R$ ($S_L$), is shown as a colored region bounded by a dotted green (red) line, consisting of all policies that are more (less) likely to reach $R$ than $L$ by a probability margin of $\epsilon$ or more. Policies that are roughly equally likely to reach $R$ or $L$, i.e., whose difference in $\Delta P$ is smaller than $\epsilon$, constitute an additional ``default'' telic state, $S_0$ (teal background), in which the agent is agnostic to which region is it more likely to reach. 
\begin{equation}
\label{eq:telic_states_def}
\begin{split}
&S_R=\left\{(\mu,\sigma)|\Delta P(\mu,\sigma)\geq \epsilon\right\},\\
&S_L=\left\{(\mu,\sigma)|\Delta P(\mu,\sigma)\leq -\epsilon\right\},\\
&S_0=\left\{(\mu,\sigma)||\Delta P(\mu,\sigma)|\leq \epsilon\right\}.
\end{split}
\end{equation}
    
\begin{figure}[ht]
\centering
    \includegraphics[width=0.75\textwidth]{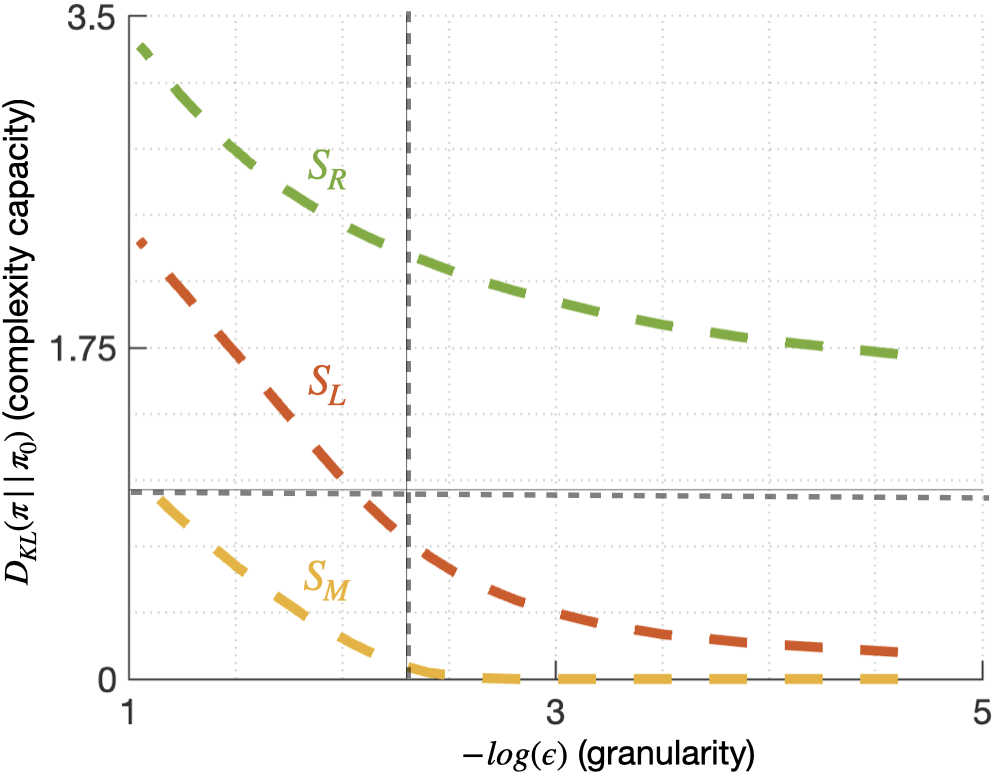} 
\caption
{\small \textbf{Complexity-granularity curves:} Each line shows the policy complexity capacity, relative to the default policy $\pi_0(0,1)$ (ordinate) required to reach the corresponding telic state at a given representational granularity level, quantified by the negative log of the sensitivity parameter $\epsilon$ (abscissa). Dashed gray lines show the values used in the dual-goal navigation example: $\delta=1$ (horizontal) and $\epsilon=0.1$ (vertical)} 
\label{fig:gc_curves}
\end{figure}

\begin{figure}[ht]
    \centering
    \includegraphics[width=0.75\textwidth]{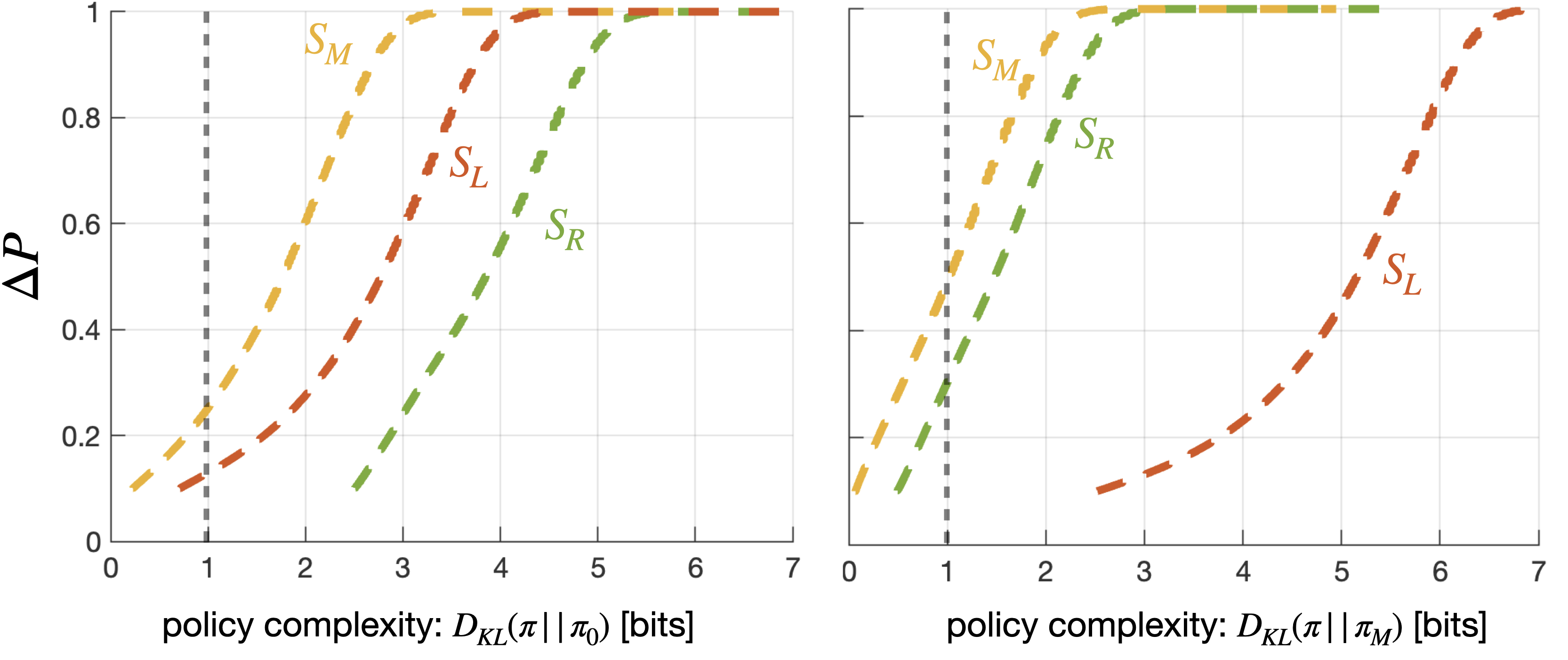}
    \caption{\small \textbf{Goal-complexity tradeoff curves:} the probability of reaching each telic state as a function of policy complexity. \textbf{Left:} an agent with a default policy $\pi_0:(\mu_0,\sigma_0)=(0,1)$ is unable to reach telic state $S_R$ with a complexity capacity limit of $\delta=1$ (gray vertical line). \textbf{Right:} with $\pi_1:(\mu_1,\sigma_1)=(1.09,1.24)$ as its default policy, the agent can reach both $S_M$ and $S_R$ with the same policy complexity capacity.}
    \label{fig:dp_complexity_curves}
\end{figure}

Using Eqs.~\ref{eq:random_walk_final_loc_dist} and~\ref{eq:telic_states_def} we can express each telic state in closed form, for example $S_R$ can be expressed, using the standard error function, $\operatorname {erf}{(x)}=2/\sqrt{\pi}\int^x_{0} e^{-t^2}dt$, as follows:
\begin{equation*}
\label{eq:telic_state_erf}
\begin{split}
&S_R=\{(\mu,\sigma)\bigg|\frac{1}{2}\left(\operatorname {erf} {\frac {R_{1}-T\mu }{{\sqrt {2T}}\sigma }}-\operatorname {erf} {\frac {R_{2}-T\mu }{{\sqrt {2T}}\sigma }}\right)-\\
&\frac{1}{2}\left(\operatorname {erf} {\frac {L_{1}-T\mu }{{\sqrt {2T}}\sigma }}-\operatorname {erf} {\frac {L_{2}-T\mu }{{\sqrt {2T}}\sigma }}\right)\geq \epsilon \}, 
\end{split}
\end{equation*}
with similar expressions for $S_L$ and $S_0$. 
To illustrate the notion of telic-controllability (Eq.~\ref{eq:telic_controllability}) using this representation, we define the complexity, $C(\pi)$, of a policy, $\pi(\mu,\sigma)$, with respect to the agent's default policy, $\pi_0(\mu_0,\sigma_0)$, as the KL divergence, per time step, between them: 
 \begin{equation*}
C(\pi)\equiv D_{KL}(\pi\|\pi_0).
 \label{eq:policy_complexity_dkl}
 \end{equation*}
The contour lines in the first three panels of Fig.~\ref{fig:phase_plots} (top \& bottom left) show isometric policy complexity levels for an agent with a complexity capacity of $\delta=1$ bit per time step, and a default policy $\pi_0(\mu_0=0,\sigma_0=1)$. Initially, both telic states, $S_R$, and $S_L$, lie within the range of the agent's policy complexity capacity (top left). The policies in $S_R$ and $S_L$ that are closest in the $KL$ sense to $\pi_0$ (green and red dots, respectively), both lie within a range of less than $\delta$ from $\pi_0$, i.e., the state representation is telic-controllable. When the center of $R$ shifts from $x_R=2$ to $x_R=2.5$ (top right), telic state $S_R$ is no longer within complexity range $\delta$ from $\pi_0$ and the state representation becomes non-controllable. To address this (bottom left), the state representation learning algorithm described in \ref{sec:learning_algorithm}, splits $S_R$ by adding an intermediate telic state $S_M$ (orange), centered around the policy closest to $S_R$ that is still within a KL-range of $\delta$ from $\pi_0$ (yellow dot). This changes the shape of $S_R$ and $S_L$ since now the probability of reaching each of the three telic states, $S_R,S_L$ and $S_M$, is defined in with respect to the two others, e.g., $S_M=\left\{(\mu,\sigma)|\Delta P_M(\mu,\sigma)\geq \epsilon\right\}$ where $\Delta P_M=p(x_T\in M\mid \mu,\sigma)-\max\{p(x_T\in L\mid \mu,\sigma),p(x_T\in R\mid \mu,\sigma)\}$, and similarly for $S_R$ and $S_L$. Since $\pi_M$ is, by construction, within a KL range of $\delta$ from $\pi_0$, the agent can reach $S_M$ by updating its default policy to $\pi_M$ (bottom right), bringing $S_R$ into reach again. Hence, the new state representation, consisting of $S_0,S_L,S_M$ and $S_R$, is tellic-controllable. Fig.~\ref{fig:dp_complexity_curves} illustrates the telic-complexity curves, showing the probability of reaching each telic state  achievable for a given complexity capacity level (x-axis). These curves quantify the maximal gain in the probability of reaching each telic state, $S_R,S_M$ or $S_L$, relative to the other two (ordinate), for a given policy complexity capacity level, with respect to a default policy of $\pi_0$ (left) or $\pi_M$ (right) (abscissa). 
Finally, Fig.~\ref{fig:gc_curves} illustrates the granularity-complexity tradeoff: the granularity of the state representation, quantified as $-\log(\epsilon)$ (abscissa), controls the complexity capacity required to reach each state (ordinate). Finer-grained representations are generally more controllable. For a granularity level of $\epsilon=0.1$ (gray vertical line), only $S_L$ and $S_M$ are reachable from $\pi_0(0,1)$ under a complexity capacity of $\delta=1$ (gray horizontal line).

\vskip6pt

\enlargethispage{20pt}

This work was supported by grant no. U01DA050647 from the National Institute on Drug Abuse and ZIAMH002983 from the Intramural Research Program of the National Institute of Mental Health.

We wish to thank the reviewers for their thoughtful comments and Yael Niv for her feedback and support.


\vskip2pc

\bibliography{sample}

\end{document}